\documentclass[10pt,twocolumn,letterpaper]{article}

\usepackage[pagenumbers]{cvpr} % To force page numbers, e.g. for an arXiv version

\definecolor{cvprblue}{rgb}{0.21,0.49,0.74}
\usepackage[pagebackref,breaklinks,colorlinks,allcolors=cvprblue]{hyperref}

\usepackage{graphicx}
\usepackage{amsmath}
\usepackage{amssymb}
\usepackage{booktabs}
\usepackage{multicol}
\usepackage{multirow}
\usepackage{pifont}

\usepackage{float}

\usepackage{xcolor}
\usepackage{arydshln}
\usepackage{colortbl}

\definecolor{cssgreen}{rgb}{0.0, 0.5, 0.0}
\definecolor{cssred}{rgb}{1, 0, 0.0}
\definecolor{DarkGreen}{rgb}{0.43, 0.68, 0.28}

\usepackage{amssymb} % 或者使用 \usepackage{amsfonts}

\usepackage[capitalize]{cleveref}
\crefname{section}{Sec.}{Secs.}
\Crefname{section}{Section}{Sections}
\Crefname{table}{Table}{Tables}
\crefname{table}{Tab.}{Tabs.}

\usepackage{colortbl}
\usepackage{booktabs}
\usepackage{multirow}
\definecolor{greenx}{HTML}{229954}
\definecolor{aliceblue}{rgb}{0.94, 0.97, 1.0}
\definecolor{f2ecde}{HTML}{f2ecde}
\definecolor{alizarin}{rgb}{0.82, 0.1, 0.26}
\definecolor{royalblue}{RGB}{65,105,225}

\usepackage{adjustbox}
\usepackage{marvosym}
\usepackage{tikz}

\newcommand{\ourmethod}{DEIM}
\newcommand{\ourclsloss}{MAL}
\makeatletter

\title{\ourmethod: DETR with Improved Matching for Fast Convergence}

\author{Shihua Huang\textsuperscript{1}$^\dag$ \quad Zhichao Lu\textsuperscript{2} \quad Xiaodong Cun\textsuperscript{3} \quad Yongjun Yu\textsuperscript{1} \quad Xiao Zhou\textsuperscript{4} \quad Xi Shen\textsuperscript{1\Letter} \quad \\
\vspace*{1mm}
\small \textsuperscript{1}Intellindust AI Lab \quad
\small \textsuperscript{2}City University of Hong Kong \quad
\small\textsuperscript{3}Great Bay University \quad
\small\textsuperscript{4}Hefei Normal University \\
\small \emph{\Letter\ Corresponding author: shenxiluc@gmail.com; $^\dag$ Project lead.}
}

\begin{document}
% \maketitle

\setcounter{figure}{-1}

\twocolumn[{
\maketitle
\begin{center}
\vspace{-0.2cm}
    \captionsetup{type=figure}
    \hfill
    \begin{subfigure}{0.494\textwidth}
\includegraphics[width=0.97\textwidth]{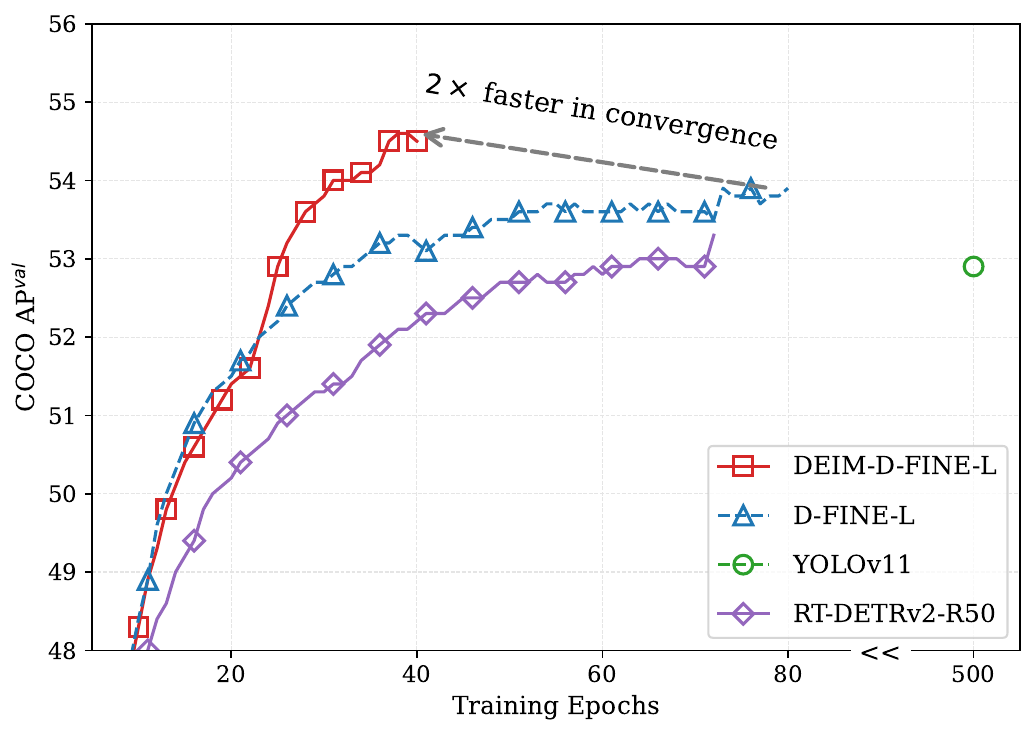}
        \caption{\textbf{Faster}: training is more compute-efficient}
        % \label{fig_train_cost}
    \end{subfigure}
    \hfill
    \begin{subfigure}{0.494\textwidth}
        \includegraphics[width=0.97\textwidth]{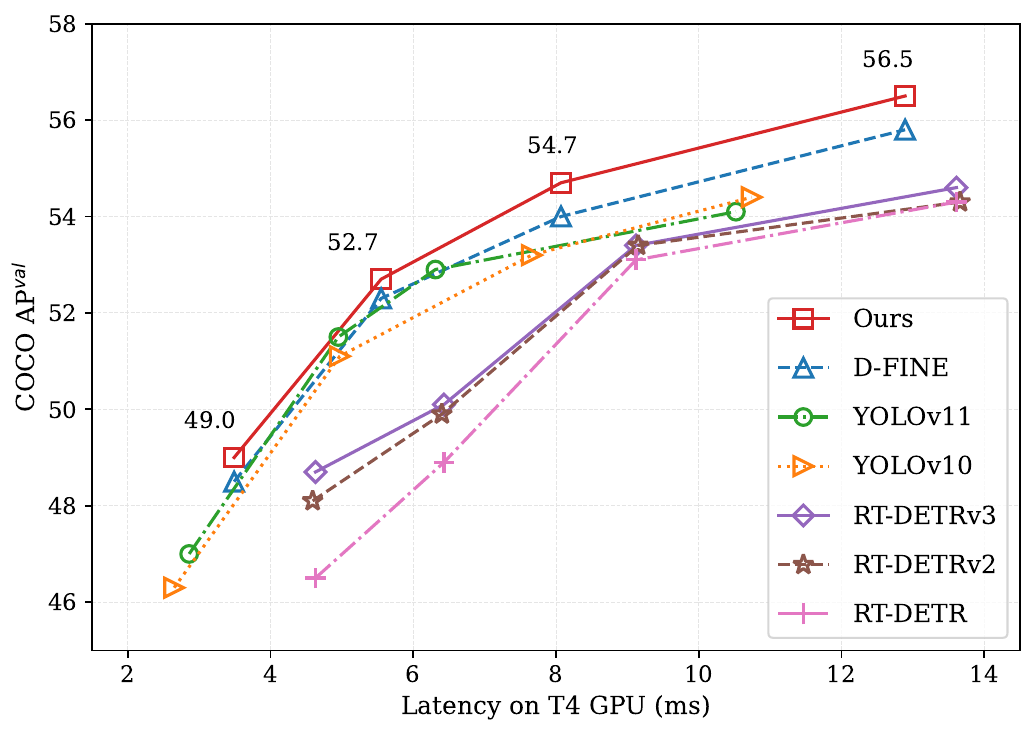}
        \caption{\textbf{Better}: exceeding all real-time detectors}
        % \label{fig_latency_ap}
    \end{subfigure}
    \hfill 
    \vspace{-0.2cm}
    \captionof{figure}{Comparison with state-of-the-art real-time object detectors on COCO~\cite{lin2014microsoft}. The proposed DEIM achieves faster convergence (a) and superior performance in terms of average precision (AP) and latency (b) when compared to state-of-the-art real-time object detectors.}
    \label{fig_front_comp}
\end{center}
}]

% \begin{figure*}[t]
%     \centering
%     \small
%     \setlength{\abovecaptionskip}{0.cm}
%     \setlength{\belowcaptionskip}{-0.cm}

%     \hfill
%     \begin{subfigure}{0.494\textwidth}
%         % \includegraphics[width=\textwidth]{figs/teaser/comp_convergence.png}
%         \includegraphics[width=0.97\textwidth]{figs/teaser/coco_epochs_vs_ap.pdf}
%         \caption{\textbf{Faster}: training is more compute-efficient}
%         \label{fig_train_cost}
%     \end{subfigure}
%     \hfill
%     \begin{subfigure}{0.494\textwidth}
%         % \includegraphics[width=\textwidth]{figs/teaser/latency_ap.png}
%         \includegraphics[width=0.97\textwidth]{figs/teaser/coco_latency_vs_ap.pdf}
%         \caption{\textbf{Better}: exceeding all real-time detectors}
%         \label{fig_latency_ap}
%     \end{subfigure}
%     \hfill 
    
%     \caption{Comparison with state-of-the-art real-time object detectors on COCO~\cite{lin2014microsoft}. The proposed DEIM achieves faster convergence (Fig.~\ref{fig_train_cost}) and superior performance in terms of average precision (AP) and latency (Fig.~\ref{fig_latency_ap}) when compared to state-of-the-art real-time object detectors.}
%     \label{fig_front_comp}
% \end{figure*}

\begin{abstract}

    We introduce DEIM, an innovative and efficient training framework designed to accelerate convergence in real-time object detection with Transformer-based architectures (DETR). To mitigate the sparse supervision inherent in one-to-one (O2O) matching in DETR models, DEIM employs a Dense O2O matching strategy. This approach increases the number of positive samples per image by incorporating additional targets, using standard data augmentation techniques. While Dense O2O matching speeds up convergence, it also introduces numerous low-quality matches that could affect performance. To address this, we propose the Matchability-Aware Loss (MAL), a novel loss function that optimizes matches across various quality levels, enhancing the effectiveness of Dense O2O.
    Extensive experiments on the COCO dataset validate the efficacy of DEIM. When integrated with RT-DETR and D-FINE, it consistently boosts performance while reducing training time by 50\%. Notably, paired with RT-DETRv2, DEIM achieves 53.2\% AP in a single day of training on an NVIDIA 4090 GPU. Additionally, DEIM-trained real-time models outperform leading real-time object detectors, with DEIM-D-FINE-L and DEIM-D-FINE-X achieving 54.7\% and 56.5\% AP at 124 and 78 FPS on an NVIDIA T4 GPU, respectively, without the need for additional data. We believe DEIM sets a new baseline for advancements in real-time object detection. 
    Our code and pre-trained models are available at \url{https://www.shihuahuang.cn/DEIM/}.
    % Our code will be made available upon publication.

\end{abstract}

\section{Introduction} \label{sec:intro}

Object detection is a fundamental task in computer vision, widely applied in fields like autonomous driving~\cite{chen2017multi, chen2016monocular}, robot navigation~\cite{ess2010object}, etc. 
The growing demand for efficient detectors has spurred the development of real-time detection methods. 
In particular, YOLO emerges as one of the main paradigms for real-time object detection, owing to its compelling trade-off between latency and accuracy~\cite{wang2024yolov9,wang2024yolov10,zheng2021yolox, bochkovskiy2020yolov4,redmon2016you}.
YOLO models are widely recognized as one-stage detectors based on convolutional neural networks. \emph{One-to-many (O2M)} assignment strategy has been widely used in YOLO series~\cite{redmon2016you,bochkovskiy2020yolov4,zheng2021yolox,wang2024yolov9}, where each target box is associated with multiple anchors. This strategy is known to be effective, as it provides dense supervision signals, which accelerate convergence and enhance performance~\cite{zheng2021yolox}.
However, it produces multiple overlapping bounding boxes per object, requiring a hand-crafted Non-Maximum Suppression (NMS) to remove redundancies, introducing latency and instability~\cite{zhao2024detrs, wang2024yolov10}.

The advent of Transformer-based detection (DETR) paradigm~\cite{carion2020end} has attracted significant attention~\cite{chen2023group, zhang2022dino, zong2023detrs}, leveraging multi-head attention to capture global context, thereby enhancing localization and classification. 
DETRs adopt a \emph{one-to-one (O2O)} matching strategy that leverages the Hungarian~\cite{kuhn1955hungarian} algorithm to establish a unique correspondence between predicted boxes and the ground-truth objects during training, eliminating the need for NMS. 
This end-to-end framework offers a compelling alternative for real-time object detection.

However, \textbf{slow convergence} remains one of the primary limitations of DETRs, and we hypothesize that the reasons are two-fold.
 % due to sparse supervision and the prevalence of low-quality matches. 
%
\ding{182} \emph{Sparse supervision}: 
The O2O matching mechanism assigns only one positive sample per target, greatly limiting the number of positive samples. In contrast, O2M generates several times more positive samples. This scarcity of positive samples restricts dense supervision, which impedes effective model learning—particularly for small objects, where dense supervision is crucial for performance.
\ding{183} \emph{Low-quality matches}: 
Unlike traditional methods that rely on dense anchors (usually $>8000$), DETR employs a small number (100 or 300) of randomly initialized queries. These queries lack spatial alignment with targets, leading to numerous low-quality matches in the training, where matched boxes have low IoU with the targets but high confidence scores.

To address the scarcity of supervision in DETR, recent studies have relaxed the constraints of O2O matching by incorporating O2M assignments into O2O training, thereby introducing auxiliary positive samples per target to increase supervision. Group DETR~\cite{chen2023group} achieves this by using multiple query groups, each with independent O2O matching, while Co-DETR~\cite{zong2023detrs} incorporates O2M methods from object detectors like Faster R-CNN~\cite{ren2016faster} and FCOS~\cite{tian2022fully}. Although these approaches successfully increase the number of positive samples, they also require additional decoders, which increases computational overhead and risks generating redundant high-quality predictions as traditional detectors.
In contrast, we propose a novel yet straightforward approach named dense one-to-one (Dense O2O) matching. Our key idea is to increase the number of targets in each training image, which in turn generates more positive samples during the training. Notably, this can be easily achieved using classical techniques such as mosaic~\cite{bochkovskiy2020yolov4} and mixup~\cite{zhang2017mixup} augmentations, which generates additional positive samples per image while preserving the one-to-one matching framework. Dense O2O matching can provide a level of supervision comparable to O2M approaches, without the added complexity and overhead typically associated with O2M methods.

% 逻辑是：sparse queries导致
Despite attempts to improve query initialization using priors~\cite{zhu2020deformable, li2022dn, zhang2022dino, zhao2024detrs}, which enable more effective query distributions around objects. 
These improved initialization methods, often relying on limited feature information extracted from the encoder~\cite{zhao2024detrs, zhang2022dino}, tend to cluster queries around a few prominent objects. In contrast, most non-salient objects lack nearby queries, leading to low-quality matches. 
This issue becomes even more pronounced when using Dense O2O. As the number of targets increases, the disparity between prominent and non-prominent targets grows, leading to a rise in low-quality matches despite the overall increase in matching quantity. In this case, if the loss function has limitations in handling these low-quality matches, this disparity will persist, hindering the model from achieving better performance.

Existing loss functions~\cite{lin2017focal, zhang2021varifocalnet} in DETRs, such as Varifocal Loss (VFL)~\cite{zhang2021varifocalnet}, are tailored to dense anchors where the number of low-quality matches is relatively low. They primarily penalize high-quality matches, especially matches with high IoU but low confidence, and discard low-quality matches. To address low-quality matches and further improve Dense O2O, we propose Matchability-Aware Loss (MAL). MAL scales the penalty based on matchability by incorporating the IoU between matched queries and targets with classification confidence. 
MAL performs similarly to VFL for high-quality matches but places greater emphasis on low-quality matches, improving the utility of limited positive samples during training. Furthermore, MAL provides a simpler mathematical formulation than VFL.
% , reducing complexity by eliminating a hyperparameter.

The proposed \ourmethod{} combines Dense O2O with \ourclsloss{} to create an effective training framework.
We conducted extensive experiments on the COCO~\cite{lin2014microsoft} dataset to evaluate the effectiveness of \ourmethod{}. The results in Fig.~\ref{fig_front_comp}~(a) show that DEIM significantly accelerates the convergence of RT-DETRv2~\cite{lv2024rt} and D-FINE~\cite{peng2024d} and achieves improved performance as well. Specifically, with only half the number of training epochs, our method outperforms RT-DETRv2 and D-FINE by 0.2 and 0.6 AP, respectively. Additionally, our approach enables training a ResNet50-based DETR model on a single 4090 GPU, achieving 53.2\% mAP within a single day (approximately 24 epochs). By incorporating more efficient models, we also introduce a new set of real-time detectors that outperform existing models, including the latest YOLOv11~\cite{yolo11}, setting a new state-of-the-art (SoTA) for real-time object detection (Fig.~\ref{fig_front_comp}~(b)).

\begin{figure*}[t]
    % \vspace{-0.2cm}
    \centering
    \small  %
    \setlength{\abovecaptionskip}{0.cm}
    \setlength{\belowcaptionskip}{-0.cm}
    \hfill
    % \hspace{-0.6cm}
    \begin{subfigure}{0.33\textwidth}
        \includegraphics[width=\textwidth]{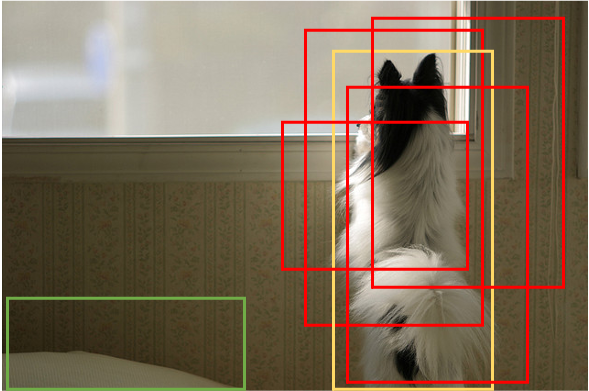}
        \caption{\textbf{O2M}: 1 target and 4 pos.}
        \label{fig:toy_O2M}
    \end{subfigure}
    \begin{subfigure}{0.33\textwidth}
        \includegraphics[width=\textwidth]{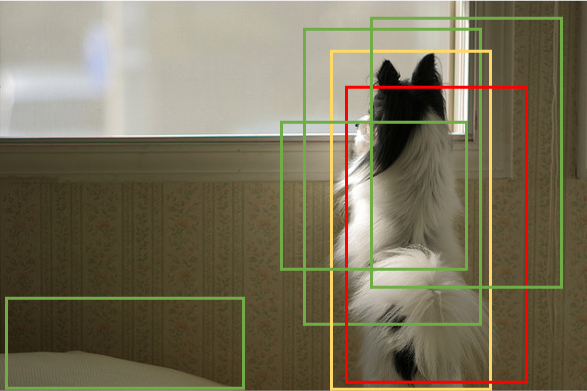}
        \caption{\textbf{O2O}: 1 target and 1 pos.}
        \label{fig:toy_O2O}
    \end{subfigure}
    \begin{subfigure}{0.328\textwidth}
        \includegraphics[width=\textwidth]{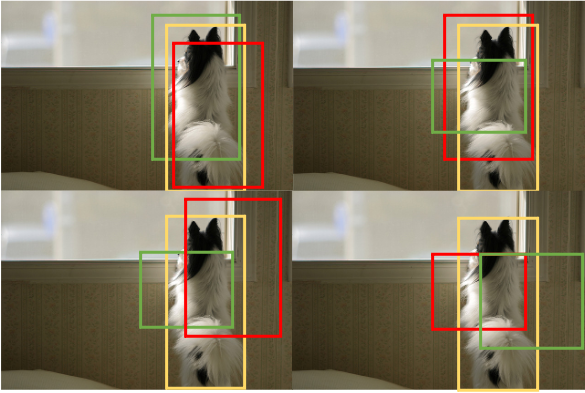}
        \caption{\textbf{Dense O2O by stitching}: 4 targets and 4 pos.}
        \label{fig:toy_Dense_O2O}
    \end{subfigure}
    % \vspace{0.1mm}
    % \hspace{-0.9cm}
    \\
    \hfill
    \begin{subfigure}{0.33\textwidth}
        \includegraphics[width=\textwidth]{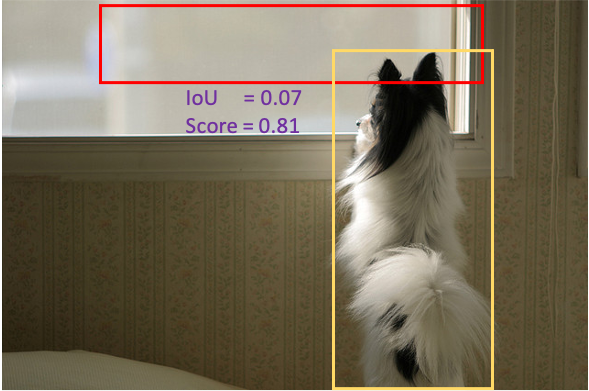}
        \caption{Low-quality matching}
        \label{fig:toy_lowquality}
    \end{subfigure}
    \hfill
    \begin{subfigure}{0.33\textwidth}
        \includegraphics[width=\textwidth, height=0.18\textheight]{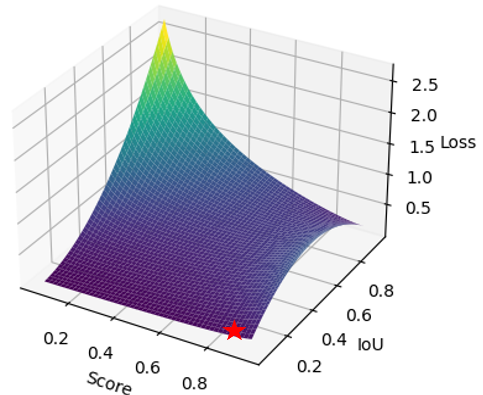}
        \caption{Loss landscape of VFL}
        \label{fig:lossscape_vfl}
    \end{subfigure}
    \begin{subfigure}{0.33\textwidth}
        \includegraphics[width=\textwidth, height=0.175\textheight]{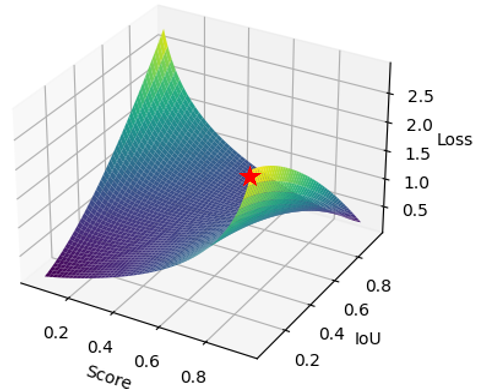}
        \caption{Loss landscape of \ourclsloss}
        \label{fig:lossscape_ours}
    \end{subfigure}
    \hfill
    \vspace{-0.2cm}
    \caption{\textbf{An illustration of our proposed \ourmethod.} \textcolor{yellow}{Yellow}, \textcolor{red}{red}, and \textcolor{green}{green} boxes represent the GT, positive and negative samples, respectively. 'pos.' denotes the positive samples. \textit{Top:} Our Dense O2O (Fig.~\ref{fig:toy_Dense_O2O}) can provide the same quality of positive samples as O2M (Fig.~\ref{fig:toy_O2M}). \textit{Bottom:} For the low-quality matching, its loss values when using VFL~\cite{zhang2021varifocalnet} and \ourclsloss\ are marked by \textcolor{red}{$\star$}, indicating \ourclsloss\ can optimize those cases more effectively.}
    \label{fig:overview}
\vspace{-0.5cm}
\end{figure*}

The main contributions of this work are summarized as follows:

\begin{itemize}

\item We introduce \ourmethod{}, a simple and flexible training framework for real-time object detection.

\item \ourmethod{} accelerates the convergence by improving the quantity and quality of matching with Dense O2O and \ourclsloss, respectively. 

\item With our method, existing real-time DETRs achieve better performance while halving training costs. Specifically, our method exceeds YOLOs and establishes a new SoTA in real-time object detection after being paired with efficient models in D-FINE. 
\end{itemize}

% The code will be made publicly available upon publication.

\section{Related Work} 
\label{sec:relate}

\textbf{Object detection with transformer (DETR)}~\cite{carion2020end} represents a shift from traditional CNN architectures to transformers. By using Hungarian~\cite{kuhn1955hungarian} algorithm for one-to-one matching, DETR eliminates the need for hand-crafted NMS as the post-processing and enables end-to-end object detection. However, it suffers from slow convergence and dense computation.

\paragraph{Increasing positive samples.}
One-to-one matching limits each target to a single positive sample, providing far less supervision than O2M and hindering the optimization. Some studies have explored ways to increase supervision within the O2O framework. Group DETR~\cite{chen2023group}, for instance, employs the concept of “groups” to approximate the O2M. It uses $K$ groups of queries, where $K > 1$, and performs O2O matching independently within each group. This allows each target to be assigned $K$ positive samples. However, to prevent communication between groups, each group requires a separate decoder layer, ultimately resulting in $K$ parallel decoders. The hybrid matching scheme in H-DETR~\cite{jia2023detrs} works similarly to Group DETR. Co-DETR~\cite{zong2023detrs} reveals that a one-to-many assignment approach helps the model learn more distinctive feature information, so it proposed a collaborative hybrid assignment scheme to enhance encoder representations through auxiliary heads with one-to-many label assignments, like Faster R-CNN~\cite{ren2016faster} and FCOS~\cite{tian2022fully}. The existing methods aim to increase the number of positive samples per target to enhance supervision. In contrast, Our Dense O2O explores another direction — increasing the number of targets per training image to boost supervision effectively. Unlike existing methods, which require additional decoders or heads and thus increase training resource consumption, our approach is computation-free.

\paragraph{Optimizing low-quality matches.} 

The sparse and randomly initialized queries lack spatial alignment with targets, resulting in a high proportion of low-quality matches that impede model convergence. Several methods have introduced prior knowledge into query initialization, such as anchor queries~\cite{wang2022anchor}, DAB-DETR~\cite{liu2022dab}, DN-DETR~\cite{li2022dn}, and dense distinct queries~\cite{zhang2023dense}. More recently, inspired by two-stage paradigms~\cite{ren2016faster, zhu2020deformable}, methods like DINO~\cite{zhang2022dino} and RT-DETR~\cite{zhao2024detrs} leverage top-ranked predictions from the encoder's dense outputs to refine decoder queries~\cite{yao2021efficient}. These strategies enable more effective query initialization closer to target regions. However, low-quality matches persist as a significant challenge~\cite{liu2023detection}. In RT-DETR~\cite{zhao2024detrs}, Varifocal Loss (VFL) is employed to reduce the uncertainty between classification confidence and box quality, enhancing real-time performance. Yet, VFL is primarily designed for traditional detectors with fewer low-quality matches and focuses on high-IoU optimization, leaving low-IoU matches under-optimized due to their minimal and flat loss values. Building on those advanced initializations, we introduce a matchability-aware loss to better optimize matches across varying quality levels, significantly enhancing the effectiveness of Dense O2O matching.

\paragraph{Reducing computation cost.}

Standard attention mechanisms involve dense computation. To improve efficiency and facilitate interactions with multi-scale features, several advanced attentions have been developed, such as deformable attention~\cite{zhu2020deformable}, multi-scale deformable attention~\cite{zhao2024ms}, dynamic attention~\cite{dai2021dynamic}, and cascade window attention~\cite{ye2023cascade}. Additionally, recent research has focused on creating more efficient encoders. For example, Lite DETR~\cite{li2023lite} introduces an encoder block that interleaves updates between high-level and low-level features, while RT-DETR~\cite{zhao2024detrs} combines CNN and self-attention in its encoder. Both designs significantly reduce resource consumption, especially RT-DETR. RT-DETR is the first real-time object detection model within the DETR framework. Building on this hybrid encoder, D-FINE~\cite{peng2024d} further optimizes RT-DETR with additional modules and refines the regression process by iteratively updating probability distributions instead of predicting fixed coordinates. This approach enables D-FINE to achieve a more favorable trade-off between latency and performance, slightly surpassing recent YOLO models. Leveraging these advancements in real-time DETRs, our method achieves impressive performance with reduced training costs, outperforming YOLO models by a substantial margin in real-time object detection.

%%%%%%%%%%%%%%%%%%%%%%%%%%%%%%%%%%%%%%%%%%%%%%%%%%%%%%%%%%%%

\section{Method}
\label{sec:method}

\subsection{Preliminaries}
\paragraph{O2M vs. O2O.} The O2M assignment strategy~\cite{zheng2021yolox, feng2021tood} is widely adopted in traditional object detectors, and its supervision can be formulated as follows:
\begin{equation} 
\text{loss} = \sum_{i=0}^N\sum_{j=0}^{M_{i}} f(\hat{y}_{ij}, y_{i}),\label{eq:assign} 
\end{equation}
where $N$ is the total number of targets, $M_{i}$ is the number of matches for the $i$-th target, $\hat{y}_{ij}$ represents the $j$-th match for the $i$-th target, $y_{i}$ denotes the $i$-th ground-truth label, and $f$ is the loss function. O2M enhances supervision by increasing $M_{i}$, i.e., assigning multiple queries to each target ($M_{i} > 1$) and thus providing dense supervision, as illustrated in Fig.~\ref{fig:toy_O2M}. In contrast, the O2O assignment only pairs each target with a single best prediction, determined via the Hungarian algorithm, which minimizes a cost function balancing classification and localization errors (Fig.~\ref{fig:toy_O2O}). O2O can be considered a special case of O2M where $M_{i} = 1$ for all targets.

\paragraph{Focal loss.}
Focal loss (FL)~\cite{lin2017focal} was introduced to prevent an abundance of easy negatives from overwhelming the detector during training, directing focus instead towards a sparse set of hard examples. It serves as the default classification loss in DETRs~\cite{zhu2020deformable, zhang2022dino} and is defined as follows:
\begin{equation}
\label{eq:fl}
\text{FL}(p, y) = 
\begin{cases} 
  -\alpha (1 - p)^\gamma \log(p)  & y = 1 \\
  -(1 - \alpha) p^\gamma \log(1 - p) & y = 0,
\end{cases}
\end{equation}
where $y \in \{0, 1\} $ specifies the ground-truth class and $p \in [0, 1] $ represents the predicted probability for the foreground class. The parameter $\gamma$ controls the balance between easy and hard samples, while $\alpha$ adjusts the weighting between foreground and background classes. In the FL, only the sample’s class and confidence are considered, with no attention given to bounding box quality, i.e., localization.

% \xiaodong{should we introduce some background and overview of the following method?}
 
% \subsection
% \noindent

\begin{figure}[t]
    \centering
    \small
    \setlength{\abovecaptionskip}{0.cm}
    \setlength{\belowcaptionskip}{-0.cm}

    \hfill
    % \hspace{-0.8cm}
    \begin{subfigure}{0.24\textwidth}
        \includegraphics[width=\textwidth]{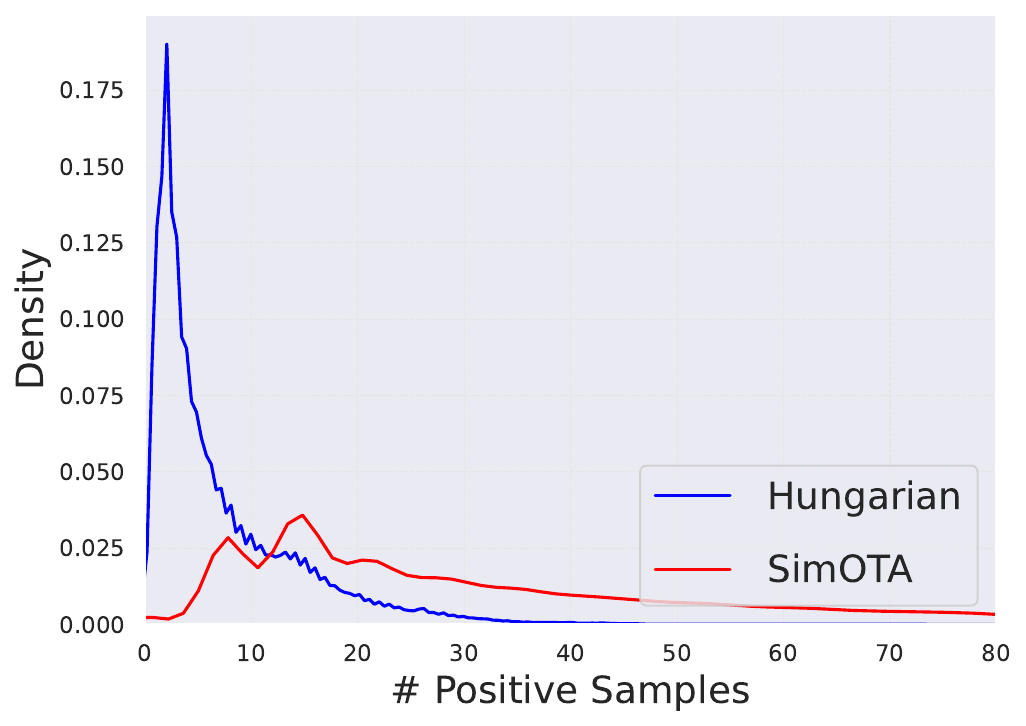}
        \caption{Matching distribution}
        \label{fig:num_match}
    \end{subfigure}
    \hfill
    \hspace{-0.2cm}
    \begin{subfigure}{0.24\textwidth}
        \includegraphics[width=\textwidth]{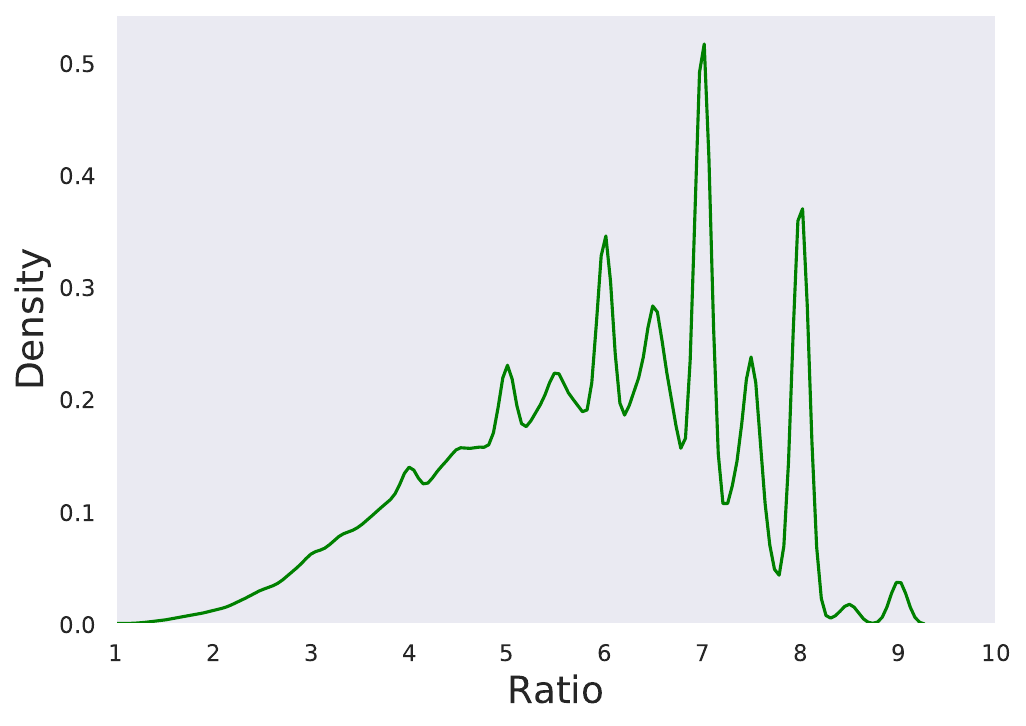}
        \caption{Ratios between O2M and O2O}
        \label{fig:ratio_match}
    \end{subfigure}
    \hfill
    \vspace{-0.2cm}
    \caption{\textbf{Anchor/Query Match Comparison}. Comparison of the number of matched anchors/queries per image in one COCO epoch using one-to-many (SimOTA~\cite{zheng2021yolox}) and one-to-one (Hungarian~\cite{carion2020end}) matching schemes.}
    \label{fig:many-vs-one}
\vspace{-0.4cm}
\end{figure}

\subsection{Improving matching efficiency: Dense O2O}

The one-to-one (O2O) matching scheme, commonly used in DETR-based models, matches each target to only one predicted query. This approach, implemented via the Hungarian algorithm~\cite{kuhn1955hungarian}, allows for end-to-end training and eliminates the need for NMS. However, a key limitation of O2O is that it generates significantly fewer positive samples compared to traditional one-to-many (O2M) methods like SimOTA~\cite{zheng2021yolox}. This leads to sparse supervision, which can slow down convergence during training.

To better understand this issue, we trained RT-DETRv2~\cite{lv2024rt} with a ResNet50 backbone on the MS COCO dataset~\cite{lin2014microsoft}. We compared the number of positive matches generated by both Hungarian (O2O) and SimOTA (O2M) strategies. As shown in Fig.~\ref{fig:num_match}, O2O produces a sharp peak under 10 positive matches per image, while O2M generates a broader distribution with many more positive matches, sometimes exceeding 80 positive samples for a single image. Fig.~\ref{fig:ratio_match} further highlights that SimOTA generates about 10 times as many matches as O2O in extreme cases. This demonstrates that O2O has fewer positive matches, potentially slowing down optimization.

We propose \textbf{Dense O2O} as an efficient alternative. This strategy retains the one-to-one matching structure of O2O (with $M_i = 1$), but increases the number of targets ($N$) per image, achieving denser supervision. For example, as shown in Fig.~\ref{fig:toy_Dense_O2O}, we replicate the original image into four quadrants and combine them into a single composite image, maintaining the original image dimensions. This increases the number of targets from 1 to 4, boosting the supervision level in Eq.~\ref{eq:assign} while keeping the matching structure unchanged. Dense O2O achieves a level of supervision comparable to O2M but without the added complexity and computational overhead.

\subsection{Improving matching quality: Matchability-Aware Loss}

\paragraph{Limitations of VFL.}  
The VariFocal Loss (VFL)~\cite{zhang2021varifocalnet}, built on the FL~\cite{lin2017focal}, has been shown to improve object detection performance, especially in DETR models~\cite{zhao2024detrs, lv2024rt, cai2023align}. VFL loss is expressed as :
\begin{equation}
\label{eq:vfl}
\small
\text{VFL}(p, q, y) = 
\begin{cases} 
  -q(q\log(p) + (1 - q)\log(1 - p)) &  q > 0 \\
  -\alpha p^\gamma \log(1 - p) & q = 0,
\end{cases}
\end{equation}
where $q$ denotes the IoU between the predicted bounding box and its target box. For foreground samples ($q>0$), the target label is set to $q$, while background samples ($q$ = 0) have a target label of 0. VFL incorporates the IoU to improve the quality of queries in DETR~\cite{zhao2024detrs}. 

However, VFL has two key limitations when optimizing low-quality matches:  
i). \textit{Low-Quality Matches}. VFL focuses mainly on high-quality matches (high IoU). For low-quality matches (low IoU), the loss remains small, preventing the model from refining predictions for low-quality boxes. For low-quality matching (with low IoU, e.g., Fig.~\ref{fig:toy_lowquality}), however, the loss remains minimal (marked by a \textcolor{red}{$\star$} in Fig.~\ref{fig:lossscape_vfl}). ii) \textit{Negative Samples}. VFL treats matches with no overlap as negative samples, which reduces the number of positive samples and limits effective training.

These issues are less problematic for traditional detectors due to their dense anchors and one-to-many assignment strategies. However, in the DETR framework, where queries are sparse and matching is more rigid, these limitations become more pronounced.

\paragraph{Matchability-Aware Loss.}  
To address these issues, we propose the Matchability-Aware Loss (\ourclsloss), which extends the benefits of VFL while mitigating its shortcomings. \ourclsloss\ incorporates the matching quality directly into the loss function, making it more sensitive to low-quality matches. The formula for \ourclsloss\ is:
\begin{equation}
\label{eq:xfl}
\small
\text{\ourclsloss}(p, q, y) = 
\begin{cases} 
  -q^\gamma \log(p) - (1 - q^\gamma)\log(1 - p) &  y=1 \\
  -p^\gamma \log(1 - p) & y = 0.
\end{cases}
\end{equation}

Compared to VFL, we introduce several small but
important changes. Specifically, the target label has been modified from $q$ to $q^\gamma$, simplifying the loss weights for positive and negative samples and removing the hyperparameter $\alpha$ used to balance positive and negative samples. This change helps to avoid the overemphasis on high-quality boxes and improves the overall training process. This can be easily seen from the loss landscape between VFL (in Fig.~\ref{fig:lossscape_vfl}) and \ourclsloss\ (in Fig.~\ref{fig:lossscape_ours}). Note that the impact of $\gamma$ is provided in Section~\ref{sec_discuss}.

\begin{figure}[t]
    \centering
    \small
    \setlength{\abovecaptionskip}{0.cm}
    \setlength{\belowcaptionskip}{-0.cm}
    \hfill
    \begin{subfigure}{0.235\textwidth}
        \includegraphics[width=\textwidth]{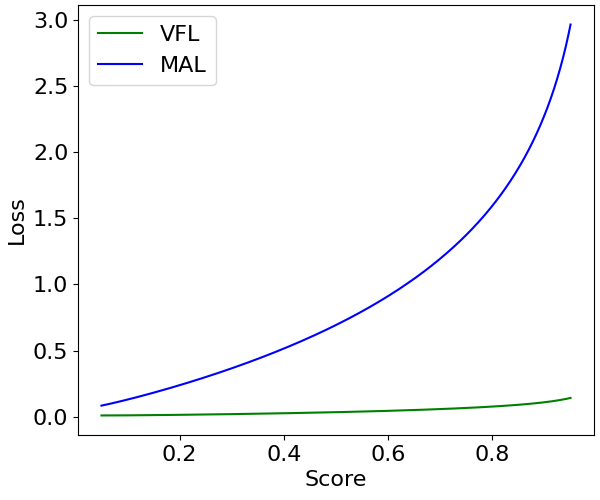}
        \caption{\textbf{Low quality}: IoU = 0.05}
        \label{fig:low_quality}
    \end{subfigure}
    \hfill 
    \begin{subfigure}{0.235\textwidth}
        \includegraphics[width=\textwidth]{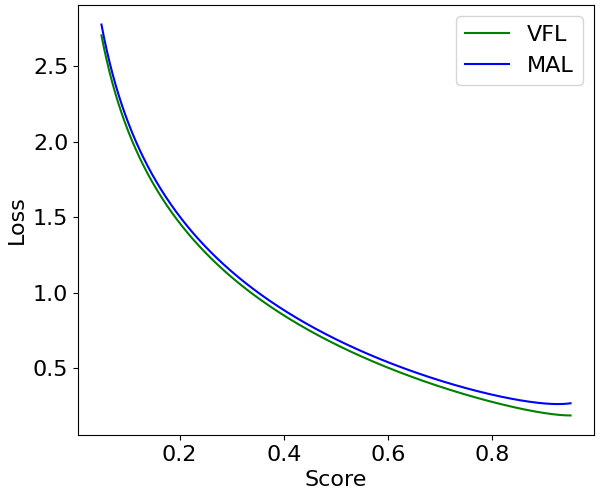}
        \caption{\textbf{High quality}: IoU = 0.95}
        \label{fig:high_quality}
    \end{subfigure}
    \hfill
    \vspace{-0.1cm}
   \caption{\textbf{VFL vs. \ourclsloss\ Comparison.} Comparison of VFL and our \ourclsloss\ for low-quality (IoU = 0.05, Fig.~\ref{fig:low_quality}) and high-quality (IoU = 0.95, Fig.~\ref{fig:high_quality}) matching cases.}
    \label{fig:loss_case}
\vspace{-0.4cm}
\end{figure}

\paragraph{Comparison with VFL.}  
We compare \ourclsloss\ and VFL in handling both low-quality and high-quality matches. In the case of low-quality matches (IoU = 0.05, in Fig.~\ref{fig:low_quality}), \ourclsloss\ shows a sharper increase in loss as predicted confidence grows, compared to VFL, which remains almost unchanged. For high-quality matches (IoU = 0.95, in Fig.~\ref{fig:high_quality}), both \ourclsloss\ and VFL perform similarly, confirming that \ourclsloss\ improves training efficiency without compromising the performance on high-quality matches.

\begin{table*}[t]
\vspace{-0.2cm}
    \caption{\textbf{Comparison with real-time object detectors on COCO~\cite{lin2014microsoft} val2017.} By integrating our method into D-FINE-L~\cite{peng2024d} and D-FINE-X~\cite{peng2024d}, we build \ourmethod-D-FINE-L and \ourmethod-D-FINE-X. We compare our method with YOLO-based and DETR-based real-time object detectors. ${\star}$ indicates that the NMS is tuned with a confidence threshold of 0.01.}
    \vspace{-0.1cm}
    \centering
    \resizebox{.94\textwidth}{!}
    {
    \begin{tabular}{l | cccc | ccc ccc}
    \toprule
    % \hline
        \textbf{Model} & \textbf{\#Epochs} & \textbf{\#Params} & \textbf{GFLOPs} & \textbf{Latency (ms)} & \textbf{AP$^{val}$} & \textbf{AP$^{val}_{50}$} & \textbf{AP$^{val}_{75}$} & \textbf{AP$^{val}_S$} & \textbf{AP$^{val}_M$} & \textbf{AP$^{val}_L$} \\
    \midrule
    % \hline
    \rowcolor{f2ecde}
    \multicolumn{11}{c}{\textbf{YOLO-based Real-time Object Detectors}} \\
    YOLOv8-L~\cite{yolov8} & 500 & 43 & 165  & 12.31 & 52.9 & 69.8 & 57.5 & 35.3 & 58.3 & 69.8 \\
    YOLOv8-X~\cite{yolov8} & 500 & 68 & 257  & 16.59 & 53.9 & 71.0 & 58.7 & 35.7 & 59.3 & 70.7 \\
    YOLOv9-C~\cite{wang2024yolov9} & 500 & 25 & 102 & 10.66 & 53.0 & 70.2 & 57.8 & 36.2 & 58.5 & 69.3 \\
    YOLOv9-E~\cite{wang2024yolov9} & 500 & 57 & 189 & 20.53 & 55.6 & 72.8 & 60.6 & 40.2 & 61.0 & 71.4 \\
    Gold-YOLO-L~\cite{wang2024gold} & 300 & 75 & 152 & 9.21 & 53.3 & 70.9 & - & 33.8 & 58.9 & 69.9 \\
    % RTMDet-L & 52M & 80 & 14.23 & 51.3 & 68.9 & 55.9 & 33.0 & 55.9 & 68.4 \\
    % RTMDet-X & 95M & 142 & 21.59 & 52.8 & 70.4 & 57.2 & 35.9 & 57.3 & 69.1 \\
    YOLOv10-L$^{\star}$~\cite{wang2024yolov10} & 500 & 24 & 120 & 7.66 & 53.2 & 70.1 & 58.1 & 35.8 & 58.5 & 69.4 \\
    YOLOv10-X$^{\star}$~\cite{wang2024yolov10} & 500 & 30 & 160 & 10.74 & 54.4 & 71.3 & 59.3 & 37.0 & 59.8 & 70.9 \\
    YOLO11-L$^{\star}$~\cite{yolo11} & 500 & 25 & 87 & 6.31 & 52.9 & 69.4 & 57.7 & 35.2 & 58.7 & 68.8 \\
    YOLO11-X$^{\star}$~\cite{yolo11} & 500 & 57 & 195 & 10.52 & 54.1 & 70.8 & 58.9 & 37.0 & 59.2 & 69.7 \\ 
    \midrule
    % \hline
    \rowcolor{f2ecde}
    \multicolumn{11}{c}{\textbf{DETR-based Real-time Object Detectors}} \\
    RT-DETR-HG-L~\cite{zhao2024detrs} & 72 & 32 & 107 & 8.77 & 53.0 & 71.7 & 57.3 & 34.6 & 57.4 & 71.2 \\ 
    RT-DETR-HG-X~\cite{zhao2024detrs} & 72 & 67 & 234 & 13.51 & 54.8 & 73.1 & 59.4 & 35.7 & 59.6 & 72.9 \\
    % \midrule
    D-FINE-L~\cite{peng2024d} & 72 & 31 & 91 & 8.07 & 54.0 & 71.6 & 58.4 & 36.5 & 58.0 & 71.9 \\
    \rowcolor[gray]{0.95} \textbf{\ourmethod-D-FINE-L} & \bf 50 & 31 & 91 & 8.07 & \textbf{54.7} & 72.4 & 59.4 & 36.9 & 59.6 & 71.8 \\ 
    D-FINE-X~\cite{peng2024d} & 72 & 62 & 202 & 12.89 & 55.8 & 73.7 & 60.2 & 37.3 & 60.5 & 73.4 \\ 
    \rowcolor[gray]{0.95}
    \textbf{\ourmethod-D-FINE-X} & \bf 50 & 62 & 202 & 12.89 & \textbf{56.5} & 74.0 & 61.5 & 38.8 & 61.4 & 74.2 \\
    \bottomrule
    \end{tabular}
    % \end{center}
    \label{tab:comp_rt}
}
\vspace{-0.2cm}
\end{table*}

\section{Experiments}
\label{sec:exps}

\subsection{Training details}
For Dense O2O, we apply mosaic augmentation~\cite{bochkovskiy2020yolov4} and mixup augmentation~\cite{zhang2017mixup} to generate additional positive samples per image. The impact of these augmentations is discussed in Section~\ref{sec_discuss}. We train our models on the MS-COCO dataset~\cite{lin2014microsoft} using the AdamW optimizer~\cite{loshchilov2017decoupled}. Standard data augmentations, such as color jitter and zoom-out, are used, as in RT-DETR~\cite{lv2024rt,zhao2024detrs} and D-FINE~\cite{peng2024d}. We employ a flat cosine learning rate scheduler~\cite{lyu2022rtmdet} and propose a novel data augmentation scheduler. A data augmentation warmup strategy is used in the first few training epochs (four usually) for simplifying attention learning. Disabling Dense O2O after 50\% of training epochs leads to better results. Following RT-DETRv2~\cite{zhao2024detrs}, we turn off data augmentation in the last two epochs. Our LR and DataAug schedulers are depicted specifically in Fig.~\ref{fig:sup_train_scheme}. Our backbones are pre-trained on ImageNet1k~\cite{deng2009imagenet}. We evaluate our models on the MS-COCO validation set at a resolution of $640 \times 640$. Additional details about the hyperparameters are provided in the supplementary material.

\begin{table*}[!t]
\caption{\textbf{Comparison with ResNet-based DETRs on COCO~\cite{lin2014microsoft} val2017.} By integrating our method into ResNet50~\cite{he2016deep} and ResNet101~\cite{he2016deep}, we build \ourmethod-RT-DETRv2-R50 and \ourmethod-RT-DETRv2-R101. We compare our method with competitive DETR-based object detectors that use ResNet50~\cite{he2016deep} or ResNet101~\cite{he2016deep} as backbones.}
\vspace{-0.15cm}
\footnotesize
\centering
\begin{tabular}{l | ccc | cccccc}
    \toprule
    \textbf{Model} & \textbf{\#Epochs} & \textbf{\#Params} & \textbf{GFLOPs} & \textbf{AP$^{val}$} & \textbf{AP$^{val}_{50}$} & \textbf{AP$^{val}_{75}$} & \textbf{AP$^{val}_S$} & \textbf{AP$^{val}_M$} & \textbf{AP$^{val}_L$} \\ 
    \midrule
    % \hline
    \rowcolor{f2ecde}
    \multicolumn{10}{c}{\textbf{ResNet50~\cite{he2016deep}-based}} \\
    DETR-DC5~\cite{carion2020end} & 500 & 41 &  187 & 43.3 & 63.1 & 45.9 & 22.5 & 47.3 & 61.1 \\
    Anchor-DETR-DC5~\cite{wang2022anchor} & 50 & 39 &  172 & 44.2 & 64.7 & 47.5 & 24.7 & 48.2 & 60.6 \\
    Conditional-DETR-DC5~\cite{meng2021conditional} & 108 & 44 & 195 & 45.1 & 65.4 & 48.5 & 25.3 & 49.0 & 62.2 \\
    Efficient-DETR~\cite{yao2021efficient} & 36 & 35 &  210 & 45.1 & 63.1 & 49.1 & 28.3 & 48.4 & 59.0 \\
    SMCA-DETR~\cite{gao2021fast} & 108 & 40 &  152  & 45.6 & 65.5 & 49.1 & 25.9 & 49.3 & 62.6 \\
    Deformable-DETR~\cite{zhu2020deformable} & 50 & 40 & 173 & 46.2 & 65.2 & 50.0 & 28.8 & 49.2 & 61.7 \\
    DAB-Deformable-DETR~\cite{liu2022dab} & 50 & 48 &  195 & 46.9 & 66.0 & 50.8 & 30.1 & 50.4 & 62.5 \\
    % DAB-Deformable-DETR++~\cite{liu2022dab} & 50 & 47 & - & 48.7 & 67.2 & 53.0 & 31.4 & 51.6 & 63.9 \\
    DN-Deformable-DETR~\cite{li2022dn} & 50 & 48 &  195 & 48.6 & 67.4 & 52.7 & 31.0 & 52.0 & 63.7 \\
    % DN-Deformable-DETR++~\cite{li2022dn} & 50 & 47 & - & 49.5 & 67.6 & 53.8 & 31.3 & 52.6 & 65.4 \\
    DINO-Deformable-DETR~\cite{zhang2022dino} & 36 & 47 &  279 & 50.9 & 69.0 & 55.3 & 34.6 & 54.1 & 64.6 \\
    RT-DETR~\cite{zhao2024detrs} & 72 & 42 & 136 & 53.1 & 71.3 & 57.7 & 34.8 & 58.0 & 70.0 \\ 
    RT-DETRv2~\cite{lv2024rt} & 72 & 42 & 136 & 53.4 & 71.6 & 57.4 & 36.1 & 57.9 & 70.8 \\ 
     \rowcolor[gray]{0.95}
    \textbf{\ourmethod-RT-DETRv2} & \bf 36 & 42 & 136 & 53.9 & 71.7 & 58.6 & 36.7 & 58.9 & 70.9 \\ \rowcolor[gray]{0.95}
    \textbf{\ourmethod-RT-DETRv2} & \bf 60 & 42 & 136 & \textbf{54.3} & 72.3 & 58.8 & 37.5 & 58.7 & 70.8 \\ \rowcolor[gray]{0.95}
    \midrule
    % \hline
    \rowcolor{f2ecde}
    \multicolumn{10}{c}{\textbf{ResNet101~\cite{he2016deep}-based}} \\
    DETR-DC5~\cite{carion2020end} & 500 & 60 &  253 & 44.9 & 64.7 & 47.7 & 23.7 & 49.5 & 62.3 \\
    Anchor-DETR-DC5~\cite{wang2022anchor} & 50 & - & - & 45.1 & 65.7 & 48.8 & 25.8 & 49.4 & 61.6 \\
    Conditional-DETR-DC5~\cite{meng2021conditional} & 108 & 63 & 262 & 45.9 & 66.8 & 49.5 & 27.2 & 50.3 & 63.3 \\
    Efficient-DETR~\cite{yao2021efficient} & 36 & 54 &  289 & 45.7 & 64.1 & 49.5 & 28.2 & 49.1 & 60.2 \\
    SMCA-DETR~\cite{gao2021fast} & 108 & 58 &  218 & 46.3 & 66.6 & 50.2 & 27.2 & 50.5 & 63.2 \\
    RT-DETR~\cite{zhao2024detrs}  & 72 & 76 & 259 & 54.3 & 72.7 & 58.6 & 36.0 & 58.8 & 72.1 \\
    RT-DETRv2~\cite{lv2024rt}  & 72 & 76 & 259 & 54.3 & 72.8 & 58.8 & 35.8 & 58.8 & 72.1 \\
    \rowcolor[gray]{0.95}
    \textbf{\ourmethod-RT-DETRv2} & \bf 36 & 76 & 259 & 55.2 & 73.3 & 59.9 & 37.8 & 59.6 & 72.8 \\ \rowcolor[gray]{0.95}
    \textbf{\ourmethod-RT-DETRv2} & \bf 60 & 76 & 259 & \textbf{55.5} & 73.5 & 60.3 & 37.9 & 59.9 & 73.0 \\
    \bottomrule
\end{tabular}
\vspace{-0.45cm}
\label{tab:comp_detrs}
\end{table*}

\subsection{Comparisons with real-time detectors}
We integrate our method into D-FINE-L~\cite{peng2024d} and D-FINE-X~\cite{peng2024d} building our \ourmethod-D-FINE-L and \ourmethod-D-FINE-X. We then evaluate these models and benchmark their real-time object detection performance against state-of-the-art models, including YOLOv8~\cite{yolov8}, YOLOv9~\cite{wang2024yolov9}, YOLOv10~\cite{wang2024yolov9}, YOLOv11~\cite{yolo11}, as well as DETR-based models like RT-DETRv2~\cite{lv2024rt} and D-FINE~\cite{peng2024d}. Tab.~\ref{tab:comp_rt} compares the models in terms of epochs, parameters, GFLOPs, latency, and detection accuracy. Additional comparisons of smaller model variants (S and M) are included in the supplementary material.

 Our method outperforms the current state-of-the-art models in training cost, inference latency, and detection accuracy, setting a new benchmark for real-time object detection. Note that D-FINE~\cite{peng2024d} is a very recent work that enhances the performance of RT-DETRv2~\cite{lv2024rt} by incorporating distillation and bounding box refinement, establishing itself as a leading real-time detector. Our DEIM further boosts the performance of D-FINE, achieving a 0.7 AP gain while reducing training costs by 30\%, with no added inference latency. The most significant improvement is observed in small object detection, where D-FINE-X~\cite{peng2024d}, when trained with our method, achieves a 1.5 AP gain as \ourmethod-D-FINE-X.

When compared directly to YOLOv11-X~\cite{yolo11}, our \ourmethod-D-FINE-L outperforms this SoTA model, achieving slightly higher performance (54.7 vs. 54.1 AP) and reducing inference time by 20\% (8.07 ms vs. 10.74 ms). Although YOLOv10~\cite{wang2024yolov9} uses a hybrid O2M and O2O assignment strategy, our models consistently outperform YOLOv10, demonstrating the effectiveness of Dense O2O.

Despite significant improvements in small object detection over other DETR-based models, our approach shows a slight decrease in small object AP compared to YOLO models. For example, YOLOv9-E~\cite{wang2024yolov9} outperforms D-FINE-L~\cite{peng2024d} by approximately 1.4 AP on small objects, though our model achieves a higher overall AP (56.5 vs. 55.6). This gap underscores the ongoing challenges in small object detection within the DETR architecture and suggests potential areas for further improvement.

\subsection{Comparisons with ResNet~\cite{he2016deep}-based DETRs}
Most DETR research uses ResNet~\cite{he2016deep} as the backbone, and to enable a comprehensive comparison across existing DETR variants, we also applied our method to RT-DETRv2~\cite{lv2024rt}, a state-of-the-art DETR variant. The results are summarized in Tab.~\ref{tab:comp_detrs}. Unlike the original DETR, which requires 500 epochs for effective training, recent DETR variants, including ours, reduce training time while improving model performance. Our method shows the most significant improvements, surpassing all variants after just 36 epochs. Specifically, \ourmethod\ reduces training time by half and increases AP by 0.5 and 0.9 on RT-DETRv2~\cite{lv2024rt} with ResNet-50~\cite{he2016deep} and ResNet-101~\cite{he2016deep} backbones, respectively. Moreover, it outperforms DINO-Deformable-DETR~\cite{zhang2022dino} by 2.7 AP with the ResNet-50~\cite{he2016deep} backbone.

\ourmethod\ also significantly enhances small-object detection. For example, while achieving comparable overall AP to RT-DETRv2~\cite{lv2024rt}, our \ourmethod-RT-DETRv2-R50\ surpasses RT-DETRv2 by 1.3 AP on small objects. This improvement is even more pronounced with the larger ResNet-101 backbone, where our \ourmethod-RT-DETRv2-R101\ outperforms RT-DETRv2-R101 by 2.1 AP on small objects. Extending training to 72 epochs further improves overall performance, especially with the ResNet-50 backbone, indicating that smaller models benefit from additional training.

\begin{table}[t]
\centering
\caption{\textbf{Comparison of the D-FINE and when with our \ourmethod{} on CrowdHuman~\cite{shao2018crowdhuman}.} Both are trained with 120 epochs.}
% \small
\vspace{-0.1cm}
\resizebox{.44\textwidth}{!}{%
\begin{tabular}{cccccccc}
    \toprule
    Method  & AP & AP$_{50}$ & AP$_{75}$ & AP$_{s}$ & AP$_{m}$ & AP$_{l}$ \\
    \hline
    D-FINE-L & 56.0 & 87.2 & 59.4 & 29.0 & 46.1 & 54.6 \\ 
    w/ DEIM & \bf 57.5 & \bf 87.6 & \bf 62.9 & \bf 33.2 & \bf 48.7 & \bf 55.7 \\ 
    \bottomrule 
\end{tabular}%
}
\vspace{-0.5cm}
\label{tab:crowdhuman} 
\end{table}

% For Additional Crowd-Human Dataset results
\subsection{Comparisons on CrowdHuman}
CrowdHuman~\cite{shao2018crowdhuman} is a benchmark dataset designed to evaluate object detectors in dense crowd scenarios. We applied both D-FINE and our proposed method to the CrowdHuman dataset, following the configurations provided in the D-FINE.
% official repository~\footnote{https://github.com/Peterande/D-FINE/configs/dfine/crowdhuman}. 
As shown in Tab.~\ref{tab:crowdhuman}, our approach (D-FINE-L enhanced with \ourmethod{}) achieves a notable improvement of 1.5 AP over D-FINE-L. In particular, our method delivers a significant performance boost (greater than 3\% improvement) on small objects (AP$_{s}$) and high-quality detections (AP$_{75}$), demonstrating its ability to detect objects more accurately in challenging scenarios. Furthermore, this experiment underscores the strong generalization capability of \ourmethod{} across diverse datasets, confirming its robustness.

\begin{table}[t]
\centering
\caption{\textbf{Comparison of Dense O2O methods with different combinations of mosaic and mixup augmentation strategies.} The probability values denote the likelihood of applying mosaic and mixup in each mini-batch during training.}
\vspace{-0.1cm}
% \small
\resizebox{.48\textwidth}{!}{%
\begin{tabular}{cccccccc}
    \toprule
    Mosaic Prob. & Mixup Prob.  & AP & AP$_{50}$ & AP$_{75}$ & AP$_{s}$ & AP$_{m}$ & AP$_{l}$ \\
    \hline
    \rowcolor{f2ecde}
    \multicolumn{8}{c}{\textbf{Training 12 Epochs}} \\
    0.0 & 0.0 &  49.6 & 67.1 & 53.6 & 31.3 & 54.2 & 67.8 \\ 
    0.5 & 0.0 &  \bf 50.4 & \bf 68.4 & \bf 54.5 & \bf 32.7 & 54.6 & 68.1 \\ 
    0.0 & 0.5 &  50.1 & 67.7 & 54.0 & 31.1 & 54.5 & \bf 68.7 \\ 
    0.5 & 0.5 &  \bf 50.4 & 68.1 & 54.2 & \bf 32.7 & \bf 54.7 & 68.2 \\
    \midrule 
    \rowcolor{f2ecde}
    \multicolumn{8}{c}{\textbf{Training 24 Epochs}} \\
    0.0 & 0.0 &  51.7 & 69.5 & 55.8 & 32.8 & 56.4 & 69.7 \\ 
    0.5 & 0.0 &  51.9 & 70.1 & 55.9 & \bf 34.9 & 56.1 & 69.3 \\ 
    0.0 & 0.5 &  51.5 & 69.4 & 55.5 & 33.2 & 56.3 & 69.3 \\ 
    0.5 & 0.5 &  \bf 52.5 & \bf 70.6 & \bf 56.7 & \bf 34.9 & \bf 57.1 & \bf 70.1 \\
    \bottomrule 
\end{tabular}%
}
\vspace{-0.2cm}
\label{tab:data_aug} 
\end{table}

\subsection{Analysis}
\label{sec_discuss}
In the following studies, we use RT-DETRv2~\cite{lv2024rt} paired with ResNet50~\cite{he2016deep} to conduct experiments and report the performance on MS-COCO val2017 as the default setup unless otherwise specified.

\paragraph{Methods for achieving Dense O2O.} We explored two approaches to implement Dense O2O: mosaic~\cite{bochkovskiy2020yolov4} and mixup~\cite{zhang2017mixup}. Mosaic is a data augmentation that combines four images into one, while mixup overlays two images at a random ratio. Both methods effectively increase the number of targets per image, enhancing supervision during training.

As shown in Tab.~\ref{tab:data_aug}, both mosaic and mixup lead to significant improvements after 12 epochs compared to training without target augmentation, highlighting the effectiveness of Dense O2O. Moreover, combining mosaic and mixup accelerates model convergence, further emphasizing the benefits of augmented supervision. 
We further tracked the number of positive samples per image over one training epoch, with results shown in Fig.~\ref{fig:data_augs}. Compared to traditional O2O, Dense O2O significantly increases positive samples.

Overall, Dense O2O intensifies supervision by increasing target counts per image, leading to faster model convergence. Mosaic and mixup are simple, computationally efficient techniques that achieve this goal, and their effectiveness suggests further potential for exploring other methods to augment target counts during training.

\begin{table}[t]
\centering
\caption{\textbf{Impact of $\gamma$ in \ourclsloss (Eqn.~\ref{eq:xfl}).} We report the performance on COCO~\cite{lin2014microsoft} val2017 for 24 epochs.}
\vspace{-0.2cm}
\resizebox{.30\textwidth}{!}{%
\begin{tabular}{lcccc}
    \toprule
    $\gamma$ & 1.3 & 1.5 & 1.8 & 2.0 \\
    \hline
    AP & 52.2 & \textbf{52.4} & 52.1 & 51.9 \\
    \bottomrule 
\end{tabular}%
}
\vspace{-0.2cm}
\label{tab:gammas}
\end{table}

\begin{table}[t]
\centering
\caption{\textbf{Impact of Dense O2O and \ourclsloss.} We conduct experiments with RT-DETRv2-R50~\cite{lv2024rt} and D-FINE-L~\cite{peng2024d}.}
\vspace{-0.2cm}
\resizebox{.43\textwidth}{!}{%
\begin{tabular}{lcccccc}
    \toprule
    Epochs & Dense O2O & \ourclsloss & AP & AP$_{50}$ & AP$_{75}$ \\
    \hline
    \rowcolor{f2ecde}
    \multicolumn{6}{c}{\textbf{RT-DETRv2-R50~\cite{lv2024rt}}} \\
    72  &  &   & 53.4 & 71.6 & 57.4 \\
    \hline
    \multirow{2}{*}{36} 
                        & \checkmark &  & 53.6 & 71.9 & 58.2 \\   % ToBe Done
                        & \checkmark & \checkmark & \bf 53.9 & \bf 71.7 & \bf 58.6 \\
    \midrule
    \rowcolor{f2ecde}
    \multicolumn{6}{c}{\textbf{D-FINE-L~\cite{peng2024d}}} \\
    72 &  &   & 54.0 & 71.6 & 58.4 \\
    \hline
    \multirow{2}{*}{36} 
                        & \checkmark &  & 54.2 & 72.1 & 58.9 \\
                        & \checkmark & \checkmark & \bf 54.6 & \bf 72.2 & \bf 59.5 \\
    \midrule
\end{tabular}
}
\vspace{-0.7cm}
\label{tab:ab_all} 
\end{table}

\paragraph{Impact of $\gamma$ in \ourclsloss (Eqn.~\ref{eq:xfl}). } The results in Table~\ref{tab:gammas} show the effect of different $\gamma$ values on \ourclsloss\ after 24 epochs. Based on these experiments, we empirically set $\gamma$ to 1.5, as it yields the best performance.

\paragraph{Effectiveness of Dense O2O and \ourclsloss.}
Tab.~\ref{tab:ab_all} presents the effectiveness of the two core components: Dense O2O and \ourclsloss. Dense O2O significantly accelerates model convergence, achieving performance similar to the baseline after just 36 epochs, as opposed to the 72 epochs required for the original model. When combined with \ourclsloss, our method further improves performance. This improvement is largely driven by better box quality, aligning with our goal of optimizing low-quality matches to improve high-quality box predictions. Overall, Dense O2O and \ourclsloss\ consistently deliver performance gains across both RT-DETRv2 and D-FINE, demonstrating their robustness and generalizability.

% \paragraph{Training and inference time.} 
\paragraph{Training speed.} 
We provide an efficient implementation using Mosaic with caching and Mixup within batches. Tab.~\ref{tab:ab_time} shows the one-epoch training time on a single 4090 GPU, where DEIM is almost as fast as the baseline (1.183 vs. 1.181) and requires less training time to converge (71 vs. 85 hours). This highlights that our approach improves convergence while maintaining efficiency.
% For inference time, DEIM is a training framework that optimizes the training process, thus it does not impact inference time.

\begin{table}[t]
\centering
\vspace{-0.1cm}
\caption{\textbf{Training time in GPU hours.}}
\vspace{-0.1cm}
\resizebox{.38\textwidth}{!}{%
\begin{tabular}{lcccccc}
\toprule
\textbf{Method}& Epoch & \#GPU hr & AP \\ \midrule
    RT-DETRv2-R50 & 1 & 1.181 & -  \\
    w/ DEIM & 1 & 1.183 & - \\ 
    \midrule
    RT-DETRv2-R50 & 72 & $~\sim$85 & 53.4  \\
    w/ DEIM & \bf 60 & \bf $~\sim$71 & \bf 54.3 \\ 
\bottomrule
\end{tabular}
}
\vspace{-0.2cm}
\label{tab:ab_time} 
\end{table}

\paragraph{Finetuning from Object365.} 
We directly employ the pre-trained Object365 weights from D-FINE and compare the results of fine-tuning with and without DEIM. As shown in Tab.~\ref{tab:ab_obj}, DEIM achieves better performance with fewer fine-tuning epochs. It further validates that DEIM delivers consistent gains, even when pre-trained on larger datasets.

\begin{table}[t]
\centering
\vspace{-0.1cm}
\caption{\textbf{Fine-tuned results from Object365 pre-training.}}
\vspace{-0.1cm}
\resizebox{.35\textwidth}{!}{%
\begin{tabular}{lcccccc}
\toprule
\textbf{Method}& Epoch & AP & AP$_{50}$ & AP$_{75}$ \\ \midrule
    D-FINE-X & 32 & 59.3 & \bf 76.8 & 64.6  \\
    w/ DEIM & \bf 24 & \bf 59.5 & 76.4 & \bf 65.2 \\ 
\bottomrule
\end{tabular}
}
\vspace{-0.4cm}
\label{tab:ab_obj} 
\end{table}

\section{Conclusion}
In this paper, we present \ourmethod{}, a method designed to accelerate convergence in DETR-based real-time object detectors by improving matching.
\ourmethod{} integrates Dense O2O matching, which increases the number of positive samples per image, with \ourclsloss{}, a novel loss designed to optimize matches across varying quality and specifically enhance low-quality matches. 
This combination substantially improves training efficiency, allowing \ourmethod{} to achieve superior performance in fewer epochs compared to models such as YOLOv11. \ourmethod{} demonstrates clear advantages over SoTA DETR models like RT-DETR and D-FINE, showing measurable gains in detection accuracy and training speed without compromising inference latency. These attributes establish \ourmethod{} as a highly effective solution for real-time applications, with the potential for further refinement and application across other high-performance detection tasks.

\vspace{0.5cm}

\noindent \textbf{Acknowledgements.} 
We thank Xuanlong Yu, Longfei Liu, and Haiyang Xie for their valuable discussions. This work was supported by the Horizontal Project of Hefei Normal University (No. HXXM2022236) and the Key Projects of the National Natural Science Foundation of Universities in Anhui Province (No. 2023AH051302).

\clearpage
{\small
\bibliographystyle{ieee_fullname}
\bibliography{main}
}

\clearpage
\setcounter{page}{1}
\setcounter{section}{0}
\maketitlesupplementary

%%%%%%%%%%%%%%%%%%%%%%%%%%%%%%%%%%%%%%%%%%%%%%%%%%%%%%%%%%%%%
\section{Experimental Settings}
%%%%%%%%%%%%%%%%%%%%%%%%%%%%%%%%%%%%%%%%%%%%%%%%%%%%%%%%%%%%%

\noindent \textbf{Dataset and metric.}
We evaluate our method on the COCO~\cite{lin2014microsoft} dataset, training {\ourmethod} on \texttt{train2017} and validating it on \texttt{val2017}. 
Standard COCO metrics are reported, including AP (averaged over IoU thresholds from 0.50 to 0.95 with a step size of 0.05), AP$_{50}$, AP$_{75}$, and AP at different object scales: AP$_S$, AP$_M$, and AP$_L$.

\begin{table}[ht]
\caption{\textbf{Different hyperparameters for D-FINE models trained with \ourmethod{}}.}
\vspace{-0.8cm}
\begin{center}
\begin{adjustbox}{width=\columnwidth}
\begin{tabular}{l|cccc}
\multicolumn{5}{l}{}\\
\toprule
\textbf{\textbf{D-FINE}} & \textbf{X} & \textbf{L} & \textbf{M} & \textbf{S} \\
\midrule
Base LR & 5e-4 & 5e-4 & 4e-4 & 4e-4 \\
Min LR & 2.5e-4 & 2.5e-4 & 2e-4 & 2e-4 \\
Backbone LR & 5e-6 & 2.5e-5 & 4e-5 & 2e-4 \\
Backbone MinLR & 2.5e-6 & 1.25e-5 & 2e-5 & 1e-4 \\
Weight of \ourclsloss{} & 1 & 1 & 1 & 1 \\
$\gamma$ in \ourclsloss{} & 1.5 & 1.5 & 1.5 & 1.5 \\
Freeze Backbone BN & False & False & False & False \\
Decoder Act. & SiLU & SiLU & SiLU & SiLU \\
Epochs & 50 & 50 & 90 & 120 \\
\bottomrule
\end{tabular}
\vspace{-0.4cm}
\end{adjustbox}
\end{center}
\label{tab:hyper_dfine}
\end{table}

\vspace{-0.6cm}
\begin{table}[ht]
\caption{\textbf{Different hyperparameters for RT-DETRv2 models trained with \ourmethod{}}.}
\vspace{-0.8cm}
\begin{center}
\begin{adjustbox}{width=\columnwidth}
\begin{tabular}{l|ccccc}
\multicolumn{5}{l}{}\\
\toprule
\textbf{RT-DETRv2} & \textbf{X} & \textbf{L} & \textbf{M$^{\star}$} & \textbf{M} & \textbf{S} \\
\midrule
Base LR & 2e-4 & 2e-4 & 2e-4 & 2e-4 & 2e-4 \\
Min LR & 1e-4 & 1e-4 & 1e-4 & 1e-4 & 1e-4 \\
Backbone LR & 2e-6 & 2e-5 & 2e-5 & 1e-4 & 2e-4 \\
Backbone MinLR & 1e-6 & 1e-5 & 1e-5 & 5e-5 & 1e-4 \\
Weight of \ourclsloss{} & 1 & 1 & 1 & 1 & 1 \\
$\gamma$ in \ourclsloss{} & 1.5 & 1.5 & 1.5 & 1.5 & 1.5 \\
Freeze Backbone BN & False & False & False & False & False \\
Decoder Act. & SiLU & SiLU & SiLU & SiLU & SiLU \\
Epochs & 60 & 60 & 60 & 120 & 120 \\
\bottomrule
\end{tabular}
\end{adjustbox}
\end{center}
\label{tab:hyper_rtdetrv2}
\end{table}

\vspace{-0.3cm}
\begin{figure}[!ht]
    \centering
    \includegraphics[width=0.48\textwidth]{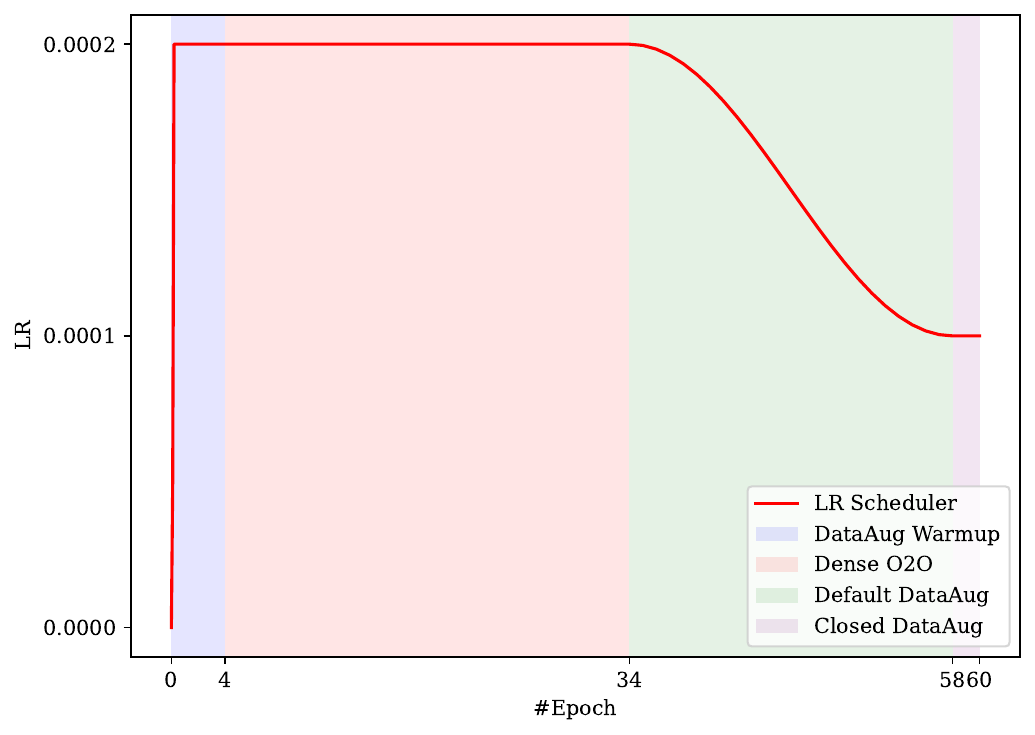}
    \vspace{-0.7cm}
    \caption{
An illustrated example of our proposed novel training scheme for learning rate and data augmentation scheduler. 
}
\vspace{-0.3cm}
\label{fig:sup_train_scheme}
\end{figure}

\noindent \textbf{Implementation details.}
We implement and validate our method using the D-FINE~\cite{peng2024d} and RT-DETRv2~\cite{zhao2024detrs,lv2024rt} frameworks. Most hyperparameters follow their original settings, with differences detailed in Tab.~\ref{tab:hyper_dfine} and Tab.~\ref{tab:hyper_rtdetrv2}, respectively. Inspired by the FlatCosine LR scheduler in RTMDet~\cite{lyu2022rtmdet}, we propose a novel data augmentation scheduler tailored for Dense O2O. Attention mechanisms in DETRs are critical for extracting accurate object features for localization and classification. However, learning attention from scratch without inductive biases can be challenging. To mitigate this, we introduce a data augmentation warmup strategy, referred to as DataAug Warmup, which simplifies the learning by disabling advanced data augmentations during the initial epochs. An example of the FlatCosine LR and proposed DataAug schedulers for 60 training epochs is shown in Fig.~\ref{fig:sup_train_scheme}.

%%%%%%%%%%%%%%%%%%%%%%%%%%%%%%%%%%%%%%%%%%%%%%%%
\section{Comparison with Lighter YOLO Detectors}
\label{sec: comparison}
%%%%%%%%%%%%%%%%%%%%%%%%%%%%%%%%%%%%%%%%%%%%%%%%
We present the results of comparisons with more lightweight real-time models (S and M sizes) in the ~\Cref{tab:model_of_SM}. Based on the strong real-time detectors RT-DETRv2~\cite{lv2024rt} and D-FINE~\cite{peng2024d}, our \ourmethod{} achieves significant improvements across the board. Notably, in RT-DETRv2, all three model sizes show an approximately 1 AP improvement, with the \ourmethod{}-RT-DETRv2-M$^{\star}$ achieving a remarkable 1.3 AP gain. Compared to other methods, our approach achieves the latest state-of-the-art results.

\begin{table*}[ht]
    \caption{\textbf{Comparison with S and M sized real-time object detectors on COCO~\cite{lin2014microsoft} \texttt{val2017}.} ${\star}$ indicates that the NMS is tuned with a confidence threshold of 0.01.}
    \vspace{-0.5cm}
    \begin{center}
    \begin{adjustbox}{width=\textwidth}
    \begin{tabular}{l | cccc | ccc ccc}
    \toprule
    \hline
        Model & \#Epochs & \#Params. & GFLOPs & Latency (ms) & AP$^{val}$ & AP$^{val}_{50}$ & AP$^{val}_{75}$ & AP$^{val}_S$ & AP$^{val}_M$ & AP$^{val}_L$ \\
    \midrule
    \hline
    \rowcolor{f2ecde}
    \multicolumn{11}{c}{\textbf{YOLO-based Real-time Object Detectors}} \\
    YOLOv8-S~\cite{yolov8} & 500 & 11 & 29 & 6.96 & 44.9 & 61.8 & 48.6 & 25.7 & 49.9 & 61.0 \\
    YOLOv8-M~\cite{yolov8} & 500 & 26 & 79 & 9.66 & 50.2 & 67.2 & 54.6 & 32.0 & 55.7 & 66.4 \\
    YOLOv9-S~\cite{wang2024yolov9} & 500 & 7 & 26 & 8.02 & 46.8 & 61.8 & 48.6 & 25.7 & 49.9 & 61.0 \\
    YOLOv9-M~\cite{wang2024yolov9} & 500 & 20 & 76 & 10.15 & 51.4 & 67.2 & 54.6 & 32.0 & 55.7 & 66.4 \\
    Gold-YOLO-S~\cite{wang2024gold} & 300 & 22 & 46 & 2.01 & 46.4 & 63.4 & - & 25.3 & 51.3 & 63.6 \\
    Gold-YOLO-M~\cite{wang2024gold} & 300 & 41 & 88 & 3.21 & 51.1 & 68.5 & - & 32.3 & 56.1 & 68.6 \\
    YOLOv10-S~\cite{wang2024yolov10} & 500 & 7 & 22 & 2.65 & 46.3 & 63.0 & 50.4 & 26.8 & 51.0 & 63.8 \\
    YOLOv10-M~\cite{wang2024yolov10} & 500 & 15 & 59 & 4.97 & 51.1 & 68.1 & 55.8 & 33.8 & 56.5 & 67.0 \\
    YOLO11-S$^{\star}$~\cite{yolo11} & 500 & 9 & 22 & 2.86 & 47.0 & 63.9 & 50.7 & 29.0 & 51.7 & 64.4 \\
    YOLO11-M$^{\star}$~\cite{yolo11} & 500 & 20 & 68 & 4.95 & 51.5 & 68.5 & 55.7 & 33.4 & 57.1 & 67.9 \\ 
    \midrule
    \hline
    \rowcolor{f2ecde}
    \multicolumn{11}{c}{\textbf{DETR-based Real-time Object Detectors}} \\
    RT-DETR-R18~\cite{zhao2024detrs} & 72 & 20 & 61 & 4.63 & 46.5 & 63.8 & 50.4 & 28.4 & 49.8 & 63.0 \\
    RT-DETR-R34~\cite{zhao2024detrs} & 72 & 31 & 93 & 6.43 & 48.9 & 66.8 & 52.9 & 30.6 & 52.4 & 66.3\\
    RT-DETRv2-S~\cite{lv2024rt} & 120 & 20 & 60 & 4.59 & 48.1 & 65.1 & 57.4 & 36.1 & 57.9 & 70.8 \\ 
    \rowcolor[gray]{0.95}  \textbf{\ourmethod-RT-DETRv2-S} & 120 & 20 & 60 & 4.59 & 49.0 & 66.1 & 53.3 & 32.6 & 52.5 & 64.1 \\ 
    RT-DETRv2-M~\cite{lv2024rt} & 120 & 31 & 92 & 6.40 & 49.9 & 67.5 & 58.6 & 35.8 & 58.6 & 72.1 \\
    \rowcolor[gray]{0.95}  \textbf{\ourmethod-RT-DETRv2-M}  & 120 & 31 & 92 & 6.40 & 50.9 & 68.6 & 55.2 & 34.3 & 54.4 & 67.1 \\ 
    RT-DETRv2-M$^*$~\cite{lv2024rt} & 72 & 33 & 100 & 6.90 & 51.9 & 69.9 & 56.5 & 33.5 & 56.8 & 69.2 \\
    \rowcolor[gray]{0.95} \textbf{\ourmethod-RT-DETRv2-M$^*$} & 60 & 33 & 100 & 6.90 & 53.2 & 71.2 & 57.8 & 35.3 & 57.6 & 70.2 \\   
    D-FINE-Nano~\cite{peng2024d} & 148 & 4 & 7 & 2.12 & 42.8 & 60.3 & 45.5 & 22.9 & 46.8 & 62.1 \\ 
    \rowcolor[gray]{0.95} \textbf{{\ourmethod}-D-FINE-Nano} & 148 & 4 & 7 & 2.12 & 43.0 & 60.4 & 46.2 & 24.5 & 47.1 & 62.1 \\ 
    D-FINE-S~\cite{peng2024d} & 120 & 10 & 25 & 3.49 & 48.5 & 65.6 & 52.6 & 29.1 & 52.2 & 65.4 \\ 
    \rowcolor[gray]{0.95} \textbf{{\ourmethod}-D-FINE-S} & 120 & 10 & 25 & 3.49 & 49.0 & 65.9 & 53.1 & 30.4 & 52.6 & 65.7 \\ 
    D-FINE-M~\cite{peng2024d} & 120 & 19 & 57 & 5.55 & 52.3 & 69.8 & 56.4 & 33.2 & 56.5 & 70.2 \\
    \rowcolor[gray]{0.95} \textbf{{\ourmethod}-D-FINE-M} & 90 & 19 & 57 & 5.55 & 52.7 & 70.0 & 57.3 & 35.3 & 56.7 & 69.5 \\
    \bottomrule
    \end{tabular}
    \end{adjustbox}
    \end{center}
    \vspace{-0.6cm}
    \label{tab:model_of_SM}
\end{table*}

\section{Additional Results}
\paragraph{Effectiveness of the minor modifications.} 
We incorporate minor modifications, including unfreezing the BN layers in the Backbone, adopting the FlatCosine LR scheduler, and replacing the Decoder activation function with SiLU, into both D-FINE-L and D-FINE-X. After training for 36 epochs, we observe that these changes have no impact on D-FINE-L but lead to a 0.1 AP improvement for D-FINE-X (55.4 vs. 55.5). This configuration is used as the new baseline for our experiments.

\begin{figure}[t]
    \centering
    \includegraphics[width=0.45\textwidth]{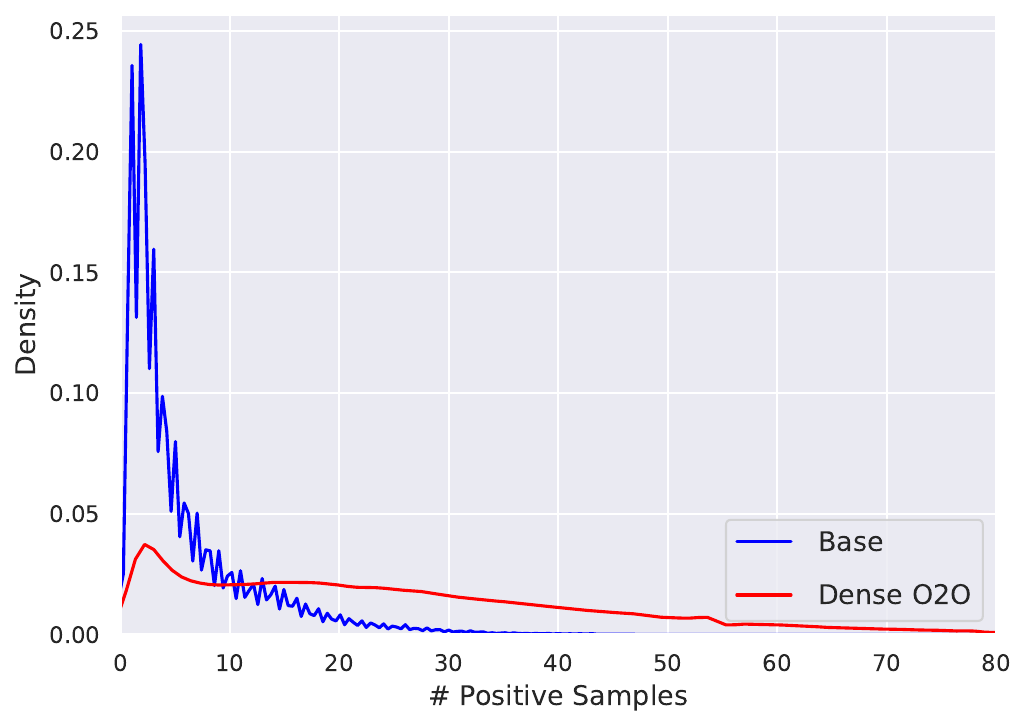}
    \vspace{-0.3cm}
    \caption{\textbf{$\#$ Positive Samples with and without Dense O2O in One Epoch of Training.} \textit{Base} indicates without Dense O2O.}
\label{fig:data_augs}
\vspace{-0.4cm}
\end{figure}

\paragraph{Number of positive samples between with/without Dense O2O.} 
During one epoch of training, we compared the number of positive samples in the same training images with and without using Dense O2O, as shown in Fig.~\ref{fig:data_augs}. After incorporating Dense O2O, the number of positive samples significantly increases. This further supports our claim that Dense O2O effectively enhances supervision.

\begin{table}[htbp]
\centering
\caption{\textbf{Varying the number of objects per training image.}}
\resizebox{.35\textwidth}{!}{
\begin{tabular}{lccccc}
\toprule
    Avg $\#$ objects & AP & AP$_{50}$ & AP$_{75}$ \\ \midrule
    \multicolumn{4}{c}{\textbf{Training 24 Epochs}} \\
    $\sim$ 10 & 51.7 & 69.5 & 55.8  \\
    \rowcolor{lightgray!50}
    $\sim$ 25 & 52.5 & 70.6 & 56.7 \\
    $\sim$ 50 & 52.2 & 70.1 & 56.4 \\
    \bottomrule
\end{tabular}
}
\label{tab:ab_number} 
\end{table}

\begin{table}[htbp]
\centering
\caption{\textbf{Training and validation accuracy.}}
\resizebox{.35\textwidth}{!}{%
\begin{tabular}{lccccc}
\toprule
    Model & AP$_{train}$ & AP$_{val}$ \\ \midrule
    RT-DETRv2-R50 & \bf 65.1 & 53.4  \\
    w/ DEIM & 64.8 & \bf 54.3 \\
    \bottomrule
\end{tabular}
}
\label{tab:ab_overfitting} 
\end{table}

\paragraph{Studies of the number of positive samples.}
We adjust the average number of objects per image during training by modifying Dense O2O. As shown in Tab.~\ref{tab:ab_number}, performance improves significantly when the number increases from 10 (without Dense O2O) to 25 (Default Dense O2O) but drops at 50 (Max Dense O2O). This decline is likely due to an imbalance in the positive-to-negative ratio and a data distribution shift caused by too many objects.
Notably, an average of 25 objects aligns with the default experimental setting used in this study, corresponding to the default Dense O2O configuration.

\paragraph{Training vs. validation accuracy.}
As shown in Tab.~\ref{tab:ab_overfitting}, DEIM achieves higher validation accuracy and slightly lower training accuracy, indicating reduced overfitting on the training set and improved adaptability to new samples.

\begin{figure}[t]
    \centering
    \includegraphics[width=0.48\textwidth]{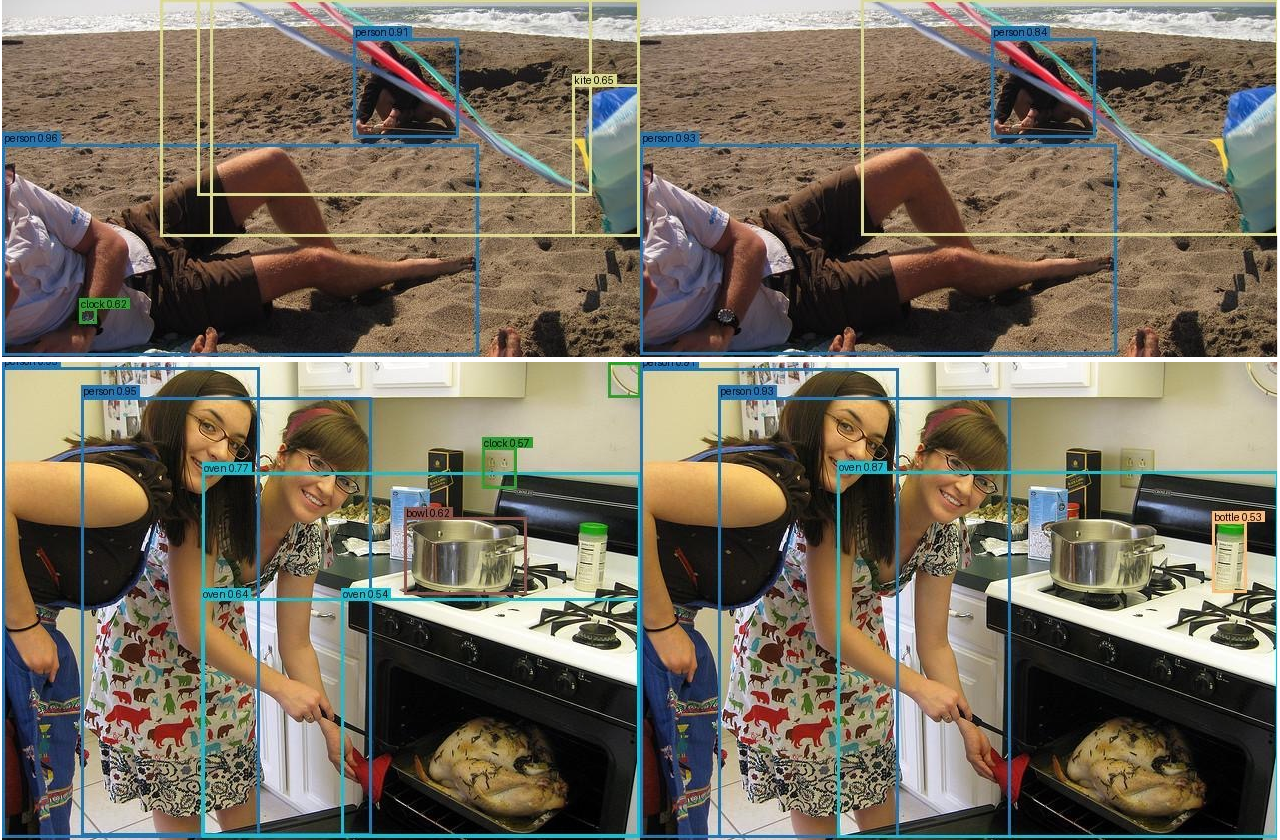}
    \vspace{-0.5cm}
    \caption{\textbf{Qualitative Comparison between D-FINE-L and \ourmethod{}.} In each paired image, the left is from D-FINE-L while the right is predicted by \ourmethod{}-D-FINE-L (Score threshold $=$ 0.5).}
\label{fig:visual}
\vspace{-0.2cm}
\end{figure}

\section{Visualizations}
We present the qualitative comparison results in Fig.~\ref{fig:visual}. These results demonstrate that \ourmethod{} effectively addresses two critical issues faced by D-FINE-L: high-confidence duplicated predictions and false positives. For example, in the top row, a single kite is erroneously assigned four highly overlapping bounding boxes, each with high confidence scores. Furthermore, as shown in the bottom row, D-FINE-L misclassifies a socket and a wall-mounted object as a clock while failing to detect the bottle. By incorporating \ourmethod{} during training, the detector successfully resolves these challenges. This visualization highlights the significant advancements enabled by \ourmethod{}, underscoring its potential for improving detection accuracy.

% {\small
% \bibliographystyle{ieee_fullname}
% \bibliography{egbib}
% }

\end{document}